\documentclass[aoas]{imsart}

\RequirePackage{amsthm,amsmath,amsfonts,amssymb}
\RequirePackage[authoryear]{natbib}
\RequirePackage{graphicx}
\usepackage{amsmath}
\usepackage{enumerate}
\usepackage{natbib}
\usepackage{url} 
\usepackage{bm}
\usepackage{amsfonts}
\usepackage{amssymb}
\usepackage[ruled]{algorithm2e}
\usepackage{xcolor}
\usepackage{epstopdf}
\usepackage{amsthm}
\usepackage{hyperref}
\hypersetup{colorlinks=true, citecolor=blue,urlcolor=blue}
\usepackage{subcaption}
\usepackage{xr,xr-hyper}
\startlocaldefs
\theoremstyle{plain}

\newtheorem{theorem}{Theorem}[section]
\newtheorem{lemma}[theorem]{Lemma}
\theoremstyle{remark}
\newtheorem{definition}[theorem]{Definition}
\newtheorem*{example}{Example}

\newcommand{\cl}{\mathcal}
\newcommand{ \tr}{\mathrm{tr}}
 
\newcommand{\wh}{\widehat}
\newcommand{\wt}{\widetilde}
\DeclareMathOperator{\vect}{vec}
\newcommand{\mbf}{\boldsymbol}
\newcommand{\bb}{\mathbb}

\newcommand{\E}{{E}}

\newcommand{\Var}{\mathrm{var}}

\newcommand{\ConvP}{\overset{P}{\rightarrow}}
\newcommand{\ConvD}{\overset{d}{\rightarrow}}

\newcommand{\qcr}{\fontfamily{qcr}\selectfont}

\DeclareMathOperator*{\argmax}{arg\,max}
\DeclareMathOperator*{\argmin}{arg\,min}

\endlocaldefs

\begin{document}

\begin{frontmatter}
\title{Optimal Sampling Designs for   Multi-dimensional Streaming Time Series with Application to Power Grid Sensor Data}
\runtitle{Optimal Sampling Designs}

\begin{aug}
\author[A]{\fnms{Rui} \snm{Xie}\ead[label=e1]{rui.xie@ucf.edu}},
\author[B]{\fnms{Shuyang} \snm{Bai}\ead[label=e2,mark]{bsy9142@uga.edu}}
\and
\author[B]{\fnms{Ping} \snm{Ma}\ead[label=e3,mark]{pingma@uga.edu}}
\address[A]{Department of Statistics and Data Science, University of Central Florida, Orlando, FL 32826 USA,
\printead{e1}}

\address[B]{Department of Statistics, University of Georgia, Athens, GA 30602 USA,
\printead{e2,e3}}
\end{aug}

\begin{abstract}
The Internet of Things (IoT) system generates massive high-speed temporally correlated 
 streaming data and is often connected with online inference tasks under computational or energy constraints.
Online analysis of these streaming time series data often faces a trade-off between statistical efficiency and computational cost. One important approach to balance this trade-off is sampling, where only a small portion of the sample is selected for the model fitting and update.
Motivated by the demands of  dynamic relationship analysis of IoT system, 
 we study the data-dependent sample selection and online inference problem for a multi-dimensional streaming time series, aiming to provide low-cost real-time analysis of high-speed power grid electricity consumption data. 
Inspired by D-optimality criterion in design of experiments, we propose a class of online data reduction methods that achieve an optimal sampling criterion and improve the computational efficiency of the online analysis.
We show that the optimal solution amounts to a strategy that is a mixture of Bernoulli sampling and leverage score sampling. 
The leverage score sampling involves auxiliary estimations that have a computational advantage over recursive least squares updates.
Theoretical properties of the auxiliary estimations involved are also discussed. 
When applied to European power grid consumption data, the proposed leverage score based sampling methods outperform the benchmark sampling method in online estimation and prediction.
The general applicability of the sampling-assisted online estimation method is assessed via simulation studies. 
 
\end{abstract}

\begin{keyword}
\kwd{Streaming data}
\kwd{Sampling design}
\kwd{Multi-dimensional time series}
\kwd{D-optimality}
\end{keyword}

\end{frontmatter}


\section{Introduction}\label{Sec:Intro}

In the era of Internet of things (IoT), the prevalence of sensor networks, wearable devices, and power grid networks   has led to   an enormous amount of  data streams being automatically collected every second, or even every millisecond.  Examples range from security monitoring in power grids
\citep{li2019online} to traffic  monitoring in smart traffic system  \citep{nellore:2016:survey}, from  health surveillance through smart wearable devices  \citep{islam:2015:internet} to soil condition sensors in precision agriculture \citep{mat:2016:iot}.
These IoT data streams carry rich and time-sensitive information on the targeted  subjects or systems, offering an unprecedented potential for real-time monitoring, forecasting and control. 
This potential, however, has not
yet been sufficiently exploited, because  the  computing
infrastructure still lags far behind the exponential growth of data sources.
For instance, a network layer of IoT system deployed in a smart power grid usually consists of a large number of low bandwidth, low energy, low processing power nodes for communication using WiFi, 3G, 4G or power line communication technologies,  rendering sophisticated  real-time analytics
infeasible \citep{jaradat2015internet}.
The IoT sensor data streams can arrive at a very high speed,  which will accumulate a large quantity of data to be analyzed in a short period of time. For real-time tasks in IoT applications, such as prediction or dynamic information flow tracking, the inference speed may lag behind the data arriving speed, especially for complex inference tasks.   
To conquer the data overflow challenge in large-scale IoT applications, we aim to provide a  sampling solution with reliable inference performance and low computation costs. 
We use publicly available power grid electricity consumption data as a motivating example to illustrate  challenges and potential solutions. 
\begin{example}
\textbf{Open Power System Data: Time series of electricity consumption}

Electricity consumption, measured by electricity loads over time in a power grid system, is a typical type of streaming data that arrives at a high speed.
In the smart IoT application, electricity consumption recorded from smart meters are capable of observing data at very high frequency, such as 25 kHz sampling frequency for the sinusoidal voltage signal~\citep{jumar2020database}.
In a power system, real-time feedback of electricity loads, which includes prediction and model fitting, is important in optimizing energy consumption patterns~\citep{marangoni2021real}.
The real-time inference helps effectively lower energy consumption by reducing energy demand and  leveling off the usage peaks.
The accurate real-time prediction will also benefit effective scheduling and decision-making in the power system~\citep{XU2021117465}.
Therefore, online analysis of the electricity load time series plays an important role in practice.

In 2020, the electricity consumption data in the United States has projected to use a total of $1,000$ million terabytes (TB) of storage~\citep{siddik2021environmental,shehabi2016united}. 
There are emerging needs for novel approaches to analyzing such massive data, especially the real-time analysis of massive data streams.
In our study, we shall work with the data streams from publicly available power system data platform \citep{realdata}, given the fact that energy data is often subject to restrictive terms of use. 
 The \cite{realdata} consists of time series of electricity consumption (load) for $37$ European countries with hourly resolution. The selected time series are recorded from 2006-01-01, 00:00:00 Coordinated Universal Time (UTC) to 2018-12-14, 23:00:00 UTC, which results in $113,544$ time points.
Electricity power consumption data from different countries are reported through different platforms in the Open Power System Data. 
We use the actual load of ENTSO-E power statistics from $19$ countries as the variables of interest. 
\begin{figure}[ht!]
	\begin{center}
		\includegraphics[width=6in]{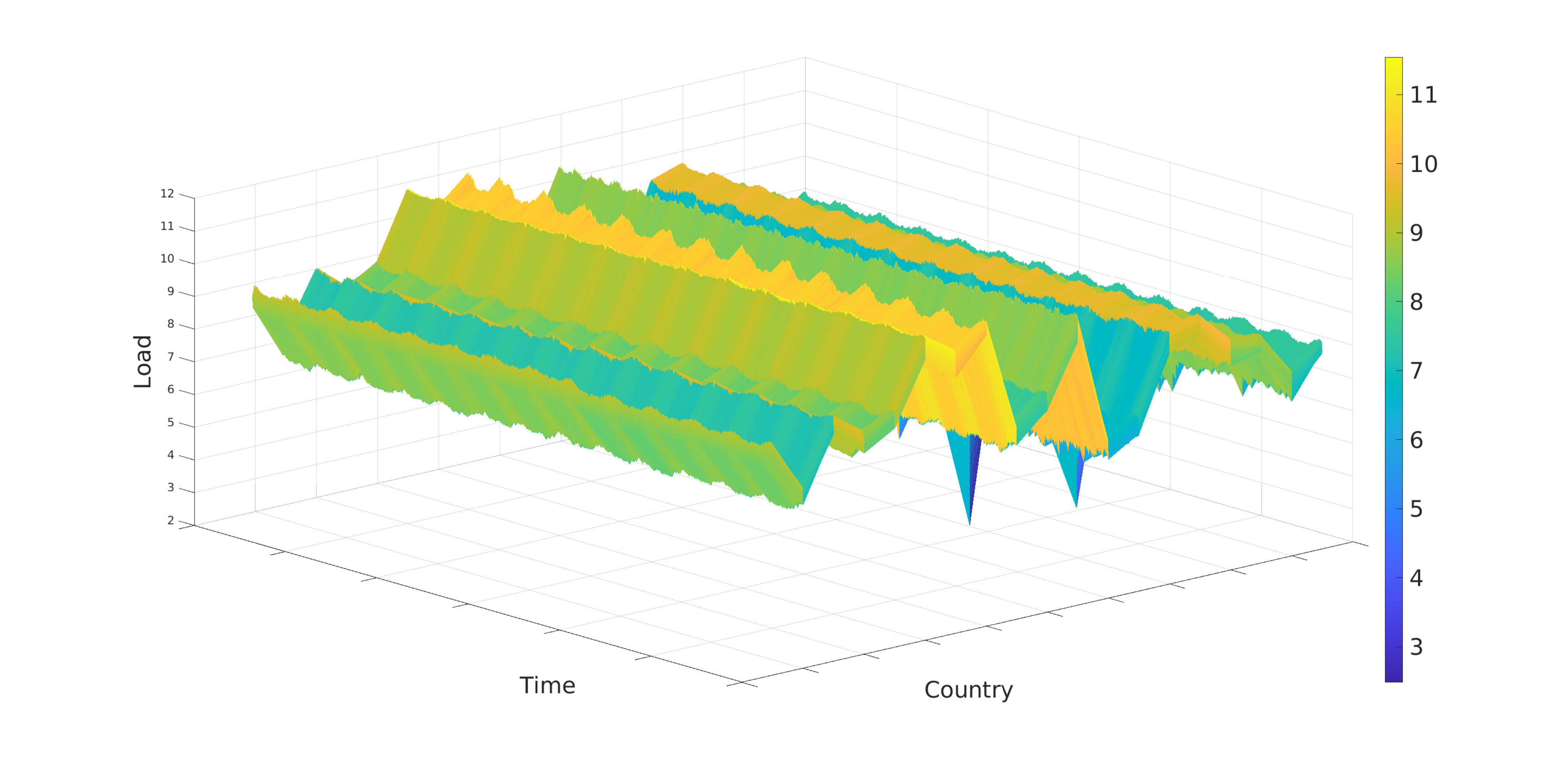}
		\caption{Load curves from 2006 to 2018 for $19$ European countries.}\label{fig:surface}
	\end{center}
\end{figure}
Figure~\ref{fig:surface} displays the dynamic evolution of the load curves from 2006 to 2018 for those $19$ countries.
\end{example}

The online analysis of the IoT systems  usually starts from  sampling~\citep{zhou2019joint} or filtering~\citep{akbar2017predictive} algorithms to optimize the data collection process from the IoT sensors, which results in massive volumes of regularly spaced multidimensional time series data streams, the latter being  a well-developed statistical subject \citep{hamilton:1994:time,lutkepohl:2005:new}.
Analyzing those large-scale high-frequency data streams posts challenges in real-time inference of multidimensional time series model, where the
computational cost is in the order of $K^3n^3$ with $n$ samples of $K$-dimensional data.
Real-time calculations, including inference and prediction, are usually critical for decision making and resource optimization in IoT applications. 

Methodologically speaking,    \emph{online analysis} of classical multidimensional time series is often carried out under the framework of dynamic linear model or state space model~\citep{west:1997:bayesian,petris:2009:dynamic}. With dynamic linear models, online statistical computations including estimation and forecasting can often be aided by the efficient Kalman filter recursive algorithms \citep{kalman:1960:new,kalman:1961:new}.  However, for applications in many IoT systems, the stringent computational resource often cannot afford to perform  typical dynamic linear model computations such as likelihood optimization or  Bayesian sampling fast enough to achieve real-time analysis.  In fact, even   restricting to linear systems, for which much less costly online estimation algorithms (e.g., recursive least squares) can be applied, a real-time inference utilizing the whole data stream can still be challenging \citep{gabel:2015:monitoring,berberidis:2016:online, hooi:2018:streamcast}.

For smart power gird analysis, due to the limited accessibility of the data, various sampling techniques, including simple random sampling and  Latin hypercube sampling~\citep{cai2013probabilistic},  are used to generate synthetic data points for Monte-Carlo simulation or static power system model training~\citep{balduin2022sampling}. The leverage score based sampling technique was used for monitoring cyberattacks in IoT system but not for the online inference~\citep{li2019online}.    To the best of our knowledge, the literature on sampling or data reduction method for online analysis of large-scale dynamic power grid systems is still lacking.
Such a situation calls for the development of new approaches.

Broadly speaking, the problem  described above belongs to a contemporary   research direction  that considers the trade-off between statistical efficiency and computational cost    \citep{jordan:2013:statistics}. 
One natural and important approach to balance  this trade-off  is    \emph{sampling}.  In this approach, a small  portion of the full observed data is chosen  as a
surrogate   to carry out computations of interest.  
Sampling, or more generally \emph{data reduction}, has been considered in various studies as a means to reduce computational cost. The majority of these studies aim to achieve certain numerical approximation accuracy, which has sparked the popularity of notions such as   sketching  (e.g., \cite{drineas2012fast,liberty:2013:simple,woodruff:2014:sketching,zhang:2017:randomization}) and coreset  (e.g., \cite{agarwal:2005:geometric, dasgupta:2009:sampling, feldman:2013:turning}).

Recently in the context of big data, some statistical investigations of sampling methods with an emphasis on   \emph{statistical estimation efficiency}   have emerged \citep{ma:2015:statistical, wang:2018:optimal, wang:2018:information,  ting:2018:optimal,yu:2020:optimal,meng2020lowcon,ma2020asymptotic,10.1214/21-AOAS1462}.  The aforementioned investigations  are all carried out under the setup of independent samples and offline computation where the whole data set is available from the beginning.  For online analysis of streaming time series exhibiting temporal dependence, relevant statistical research is still largely lacking. Exceptions are the work \cite{xie:2019:online}  that considered   online estimation of a Gaussian vector autoregression (VAR) model assisted by the so-called leverage score sampling (LSS), and the work \cite{eshragh2019lsar} that applied leverage score-based sampling for the analysis of large-scale univariate time series data. Leverage score sampling proposed in \cite{xie:2019:online}  selects samples based on thresholding a leverage score of the lagged covariate in vector autoregression. However, a number of questions regarding this sampling method  were left unaddressed.  \cite{xie:2019:online} showed that LSS achieves an asymptotic efficiency superior to a naive Bernoulli sampler. However, its optimality over other possible sampling methods was not clarified. Furthermore, the sample selection rule of LSS focuses exclusively on high-leverage covariate points, which in practice might  lead to a lack of good fit over low-leverage region and sensitivity to outliers. In addition   in  \cite{xie:2019:online}, the auxiliary estimations involved in implementing LSS were not  justified.


The overall objective of this paper is to develop methodologies for performing online  analysis on high-speed multidimensional time series, where we apply them to provide a real-time solution for massive power grid sensor  data inference in IoT applications.
On the methodology side, we propose a computationally efficient online sampling method called relaxed-LSS,  which can be applied to linear multivariate time series models including extraneous variables and enjoys more robustness compared with LSS. The relaxed-LSS and the time series model are used for online inference and prediction of the IoT power consumption data.  On the theory side, we establish the D-optimality of asymptotic estimation efficiency of LSS and more generally relaxed-LSS in a broad class of online sampling methods. We also establish consistency properties incorporating some auxiliary estimations in the sampling methods.

We organize the article as follows.
In Section \ref{Sec:linear sys}, we consider a framework that covers a large class of models subsuming the Gaussian Vector Autoregression considered in \cite{xie:2019:online}.
In Section \ref{Sec:opt sampling},  we formulate a class of online sampling methods  within the framework of Section \ref{Sec:linear sys} and show that LSS is the optimal choice among them for asymptotic estimation efficiency. We then proceed to propose a novel  relaxed version of LSS which keeps a proportion of low-leverage covariate points. 
Section \ref{Sec:aux} addresses  the  auxiliary estimations and practical implementation involved in the sampling algorithm, including  an online estimate of an inverse covariance matrix which is performed sparsely for computational advantage.
 Section \ref{Sec:data} considers  an application to the electricity consumption time series in
Open Power System Data. 
Section \ref{Sec:simulate} includes the simulation studies of LSS and its relaxed version. 
The conclusion and future work are presented in Section \ref{con}.
The proofs of the theoretical results are collected in Supplementary Material.

\section{Model Assumptions}\label{Sec:linear sys}
By default, all vectors are column vectors.  Throughout this paper,  $\|\cdot\|$ denotes  the matrix operator norm with respect to the Euclidean norm (so it is  the Euclidean norm for   vectors).

Suppose $(\mbf{y}_t)$ is the $K$-dimensional time series of interest.   For notional convenience, we allow $t$ to vary within the whole integer set $\bb{Z}$, whereas the   starting point of $t$ will become clear in a context. 
We shall  model $(\mbf{y}_t)$ as a $\bb{R}^K$-valued    ergodic (strictly) stationary process with finite variance.    Suppose, in addition, that we have a $\bb{R}^p$-valued ergodic stationary  process $(\mbf{x}_t)$ with finite variance serving as explanatory variables for $(\mbf{y}_t)$.  We impose the following stationarity assumption,  which will help simplify the formulation of the basic results.
Let $\mbf{\mu}_X=\E[\mbf{x}_t], \mbf{\mu}_Y=\E[\mbf{y}_t]$,
we shall assume the following linear system  model for the centered time series,
\begin{equation}\label{eq:stat sys}
	\mbf{y}_t-\mbf{\mu}_Y   = B'(\mbf{x}_t-\mbf{\mu}_X)       + \mbf{e}_t,
\end{equation}
where $B$ is a $p\times K$ coefficient matrix and $B'$ stands for its transpose, $(\mbf{e}_t)$ is a $\bb{R}^K$-valued ergodic stationary noise process satisfying the martingale difference property, 
\begin{equation}\label{eq:md}
	\E[\mbf{e}_t|  \cl{F}_t]=\mbf{0},
\end{equation}
where the filtration $\cl{F}_t=\sigma(\mbf{x}_s,\ \mbf{e}_{s-1},\ s\le t)$, and the conditional expectation in \eqref{eq:md} is taken component-wise. We also assume a constant conditional  covariance matrix for the error process:
\begin{equation}\label{eq:e_t cov}
	\E [\mbf{e}_{t_1} \mbf{e}_{t_2}'|\cl{F}_t]=\Omega 1_{\{t_1=t_2\}},\quad t_1,t_2\in \bb{Z}, \ t_1,t_2>t
\end{equation}
for some non-singular covariance matrix $\Omega$.     Assume,  in addition, that $(\mbf{x}_t;\mbf{e}_t)$ is jointly stationary.
In literature~(e.g., \citet[Chapter 14]{georgebox2015}), the model \eqref{eq:stat sys} is often replaced by one with the means absorbed into an intercept term in $\mbf{x}_t$. Our formulation here singles out the estimation of the means which is computationally trivial, and facilitates the development of the optimality results.

It is worth pointing out that the components of the explanatory variable process $(\mbf{x}_t)$ are allowed to contain lagged values of $(\mbf{y}_t)$. For instance, we may set
\begin{equation}\label{eq:VARX x_t}
	\mbf{x}_t = (\mbf{y}_{t-1}',\cdots,\mbf{y}_{t-p_1}', \mbf{v}_{t-1}',\cdots,\mbf{v}_{t-p_2}' )' \in \bb{R}^{K(p_1+p_2)},
\end{equation}
where $(\mbf{v}_t)$ is a  stationary process extraneous to $(\mbf{y}_t)$, where each $\mbf{v}_t \in \bb{R}^{K}$. The model \eqref{eq:stat sys} with the specification \eqref{eq:VARX x_t}  is often known as a VARX model (e.g., \citet[Chapter 10]{lutkepohl:2005:new}), which  becomes the well-known vector autoregression model if the extraneous variables are absent. The VARX model is more commonly expressed as
\begin{equation}\label{eq:VARX}
	\mbf{y}_t  = \sum_{i=1}^{p_1}\Phi_i \mbf{y}_{t-i}  +\sum_{j=1}^{p_2} \Psi_j \mbf{v}_{t-j} + \mbf{e}_t 
\end{equation} 
for some $K\times K$ coefficient matrices $\Phi_i$'s and $\Psi_i$'s, where  $\Phi_i$'s must satisfy appropriate constraints to allow the existence of stationary solution of \eqref{eq:VARX} (e.g.,~\citet[Section 10.1]{hamilton:1994:time}).

The linear system model \eqref{eq:stat sys} and its variants  have been applied in various IoT contexts for real-time analysis of streaming time series:   anomaly detection in streaming environmental sensor data \citep{hill:2010:anomaly};    tracking  causal interactions between brain
regions based on MEG sensor streams \citep{michalareas:2013:investigating};  design of energy-efficient operation of  low-power wireless medical sensors \citep{anagnostopoulos:2014:autoregressive};    traffic forecasting in large urban areas based on road network sensors \citep{schimbinschi:2017:topology}.

For the development of optimal online sampling theory in Section \ref{Sec:opt sampling}, we shall impose an additional assumption  on the distributional shape of the covariate $\mbf{x}_t$. Recall a $p$-dimensional elliptical (contoured) distribution $EC_p(\mbf{\mu},\Sigma,\nu)$ (cf.\ \cite{fang:1990:symmetric})  is specified by the following
three components: a vector
$\mbf{\mu}\in \bb{R}^p$ called  \emph{location}, a $p\times p$ non-negative definite matrix $\Sigma$ called \emph{scatter}, which coincides with the covariance matrix when the latter exists,    and  a probability distribution $\nu$ on $\bb{R}_+$. 
In particular, $EC_p(\mbf{\mu},\Sigma,\nu)$ 
denotes the distribution of random vector $
\mbf{\mu}+ \xi \Sigma^{1/2} \mbf{S}$,
where $\Sigma^{1/2}$ is a symmetric square root of $\Sigma$, the random variable $\xi$ follows the distribution   $\nu$ and is called the \emph{generating variate}, and $\mbf{S}$ is the uniform distribution on the $p$-dimensional unit sphere $\{\mbf{x}\in \bb{R}^p:\ \|\mbf{x}\|=1\}$ which is independent of $\xi$. 
A multivariate normal distribution  is included in this family as a special case.  
For the rest of the paper, we shall assume that for each fixed $t\in\bb{Z}$, the covariate vector 
\begin{equation}\label{eq:x EC}
	\mbf{x}_t\sim EC_p(\mbf{\mu}_X,\Sigma,\nu),
\end{equation}
where the distribution $\nu$ is  absolutely continuous and the scatter $\Sigma$ is also the positive definite covariance matrix of $\mbf{x}_t$. Hence the distribution of $\mbf{x}_t$ is absolutely continuous.

\section{Optimal Online Sampling}\label{Sec:opt sampling}

To develop the main ideas, throughout this section    we shall   also assume  for simplicity that
\begin{equation}\label{eq:centering}
	\mbf{\mu}_X=\mbf{\mu}_Y = \mbf{0}.
\end{equation}
In practice, online estimation of these means can be achieved  efficiently with a negligible computational cost compared to that of estimating $B$. See also Section \ref{Sec:aux} below for more details.

\subsection{A Class of Online Samplers}

When  a   sample stream $(\mbf{y}_t;\mbf{x}_t)_{t=1}^n$ from the whole stream   $(\mbf{y}_t;\mbf{x}_t)_{t=1}^\infty$ satisfying \eqref{eq:stat sys} is observed and $n\gg Kp$,   the least square estimator  
\[
\wh{B}_{n,LS}=\argmin\limits_{B}\sum_{t=1}^{n}||\mbf{y}_t - B'\mbf{x}_t ||^2
\]
may be used to estimate $B$.
In the context of online estimation, the computation of $\wh{B}_{n,LS}$ can be implemented in a recursive manner based on the Sherman–Morrison inversion formula \citep{sherman:1950:adjustment}, an algorithm often known as the \emph{recursive least squares} \citep{plackett:1950:some}. However, when  data stream arrives at an overwhelmingly fast rate with high dimension ($Kp$ is large), real-time update of $\wh{B}_{n,LS}$ can still be challenging given a limited computational capacity in the context of streaming data (see Section \ref{Sec:Intro}).

The basic idea we shall propose is simple: instead of updating the estimation along the whole stream of data   $(\mbf{y}_t;\mbf{x}_t)_{t=1}^\infty$, we shall skip some time points $t$ and only update the estimate along a subset $I\subset \bb{Z}_+$. The problem becomes: how should $I$ be selected, or how do we select samples among $(\mbf{y}_t;\mbf{x}_t)_{t=1}^\infty$?

It is useful to note that our problem bears similarity to   the study of \emph{optimal designs}   (e.g., \cite{pukelsheim:1993:optimal,papalambros:2000:principles}). Generally speaking, an optimal design aims at achieving an optimal estimation precision with a fixed sample size (number of experimental runs). 
In optimal design for linear regression with response $\mbf{y}$ and design matrix $X$, i.e, $\mbf{y}=X\mbf{\beta}+\mbf{\epsilon}$, where random error  $\mbf{\epsilon}$ has mean zero and constant variance $\sigma^2I$, one often considers optimizing the matrix $(X' X)^{-1}$ (recall  $\Var[\wh{\mbf{\beta}}]=\sigma^{2}(X' X)^{-1}$, where $\wh{\mbf{\beta}}$ is the least square estimator) based on a certain criterion, e.g., the determinant (D-optimality). 
The optimization   is typically over a finite available set of candidate covariate points (treatment runs) with the rows of $X$ (number of runs)   fixed. In other words,  given a limited sample size, one needs to select appropriate covariate points among available ones to optimize $(X' X)^{-1}$. 

Motivated by this insight drawn from   optimal design, we propose to  base our online sample selection criterion on the covariate   $\mbf{x}_t$ as well.    
We   consider the following  class of samplers: at each time point $t$, the selection of  sample unit $(\mbf{y}_t;\mbf{x}_t)$ is determined solely by  $\mbf{x}_t\in \bb{R}^p$ up to a randomization. This is made precise   as follows. 
\begin{definition}
\label{Def:S}
	Suppose there is a measurable function  $s: \mathbb{R}^{p}\rightarrow [0,1]$, called the sampling function.
	A sampling method (or sampler) $\cl{S}(s)$ is defined  as follows: conditioning on $(\mbf{y}_t;\mbf{x}_t)_{t=1}^\infty$,   for each $t\in\bb{Z}_+ $, the  sample    $(\mbf{y}_t;\mbf{x}_t)$ is selected independently with probability $s(\mbf{x}_t)$. 
	
	The sampler $\cl{S}(s)$ can be alternatively described as follows: let $(U_t)_{t=1}^\infty$  be i.i.d.\ Uniform$(0,1)$ random variables which are independent of $(\mbf{y}_t;\mbf{x}_t)_{t=1}^\infty$.  The   selected index set $I$ is given by 
	\[
	I=\{t\in \bb{Z}_+:\ U_t\le  s(\mbf{x}_t)\}.
	\]  
\end{definition} 
The (unconditional) sampling rate of $\cl{S}(s)$   is given by
\begin{equation}\label{eq:q gen}
	q:= pr(U_t\le s(\mbf{x}_t))= \E[  s(\mbf{x}_t)]\in [0,1].
\end{equation}
When a  stream of total length $n$ passes through, the   size of selected sample is given by a random number
\[
N:=\sum_{t=1}^n 1_{\{U_t\le s(\mbf{x}_t)\}}=|I|.
\]  
By the ergodic theorem \cite[Theorem 10.6]{kallenberg:2002:foundations}, we have\begin{equation}\label{eq:N/n}
	\frac{N}{n} \rightarrow  q
\end{equation}
a.s.\ as $n\rightarrow\infty$.

As an example,
a constant sampling function $s(\mbf{x})\equiv q$ corresponds to the \emph{Bernoulli sampler}, that is, each index $t$ is  independently selected  with probability $q$ regardless of $\mbf{x}_t$.

For an \emph{ideal} stationary system, one may argue that there is no need to employ the sampling-assisted approach    for online estimation.  Indeed,   if the regression coefficients in $B$ in \eqref{eq:stat sys} stays invariant along the whole  data stream, then the older data provides exactly the same   information as the newer data about $B$.   Hence one may use the available computational capacity to update the estimate up to the best speed even if it cannot catch up with the newest received data. However, in the practice of real-time analysis of streaming time series, this approach should be avoided since it fails to reflect the latest information about the streams. In fact, the usefulness of   sampling-assisted online estimation is necessarily tied to \emph{non-stationarity}, be it for the detection of departure from stationarity, or for the predictive modeling of   non-stationary  time series.  For example, for predictive modeling, one may propose a time-varying version of \eqref{eq:stat sys} and estimate  time-varying $B$ by weighted least squares (e.g.,  \cite{zhou:2010:simultaneous,zhang:2012:inference}). Since such a system can be locally viewed as stationary, the  foundation laid in this paper will still be highly relevant in that context.

\subsection{D-Optimality of Leverage Score Sampler}

The next step is to formulate optimality among the class   $\{S(s)\}$.
Note that unlike a conventional setup in optimal design,    for online estimation the number of rows of the design matrix $X$ keeps increasing. So we propose to  formulate the optimization, in a sense, on an asymptotic version of  $(X^T X)^{-1}$.  
Let $\wh{B}_{n,I}$ be the least squares estimator of $B$ using only $(\mbf{y}_t;\mbf{x}_t)_{t\in I}$, namely,
\begin{equation}\label{eq:B LS}
	\wh{B}_{n,I}= \left( \sum_{t\in I}  \mbf{x}_t   \mbf{x}_t'  \right)^{-1} \left( \sum_{t\in I} \mbf{x}_t \mbf{y}_t'   \right)=\left( \sum_{t=1}^n  1_{\{U_t\le s(\mbf{x}_t)\}} \mbf{x}_t   \mbf{x}_t'  \right)^{-1} \left( \sum_{t=1}^n 1_{\{U_t\le s(\mbf{x}_t)\}} \mbf{x}_t \mbf{y}_t'   \right).
\end{equation} 
We have  the following asymptotic normality result.
\begin{theorem}\label{Thm:asymp_normality}
	Under the assumptions in Section \ref{Sec:linear sys}, suppose in addition that 
	\begin{equation}\label{eq:Gamma(s)}
		\Gamma(s)=\E\left[ s(\mbf{x}_t) \mbf{x}_t \mbf{x}_t' \right]
	\end{equation} is invertible. Then
	as    the total stream size $n\rightarrow\infty$,  
	\begin{equation}\label{eq:asymp norm lev} 
		\sqrt{N}(\vect(\wh{B}_{n,I})  - \vect(B))\sim  \sqrt{nq}(\vect(\wh{B}_{n,I})  - \vect(B))\ConvD \cl{N}(\mbf{0},  P(s)^{-1} ),
	\end{equation}
	where     $\vect(B)$ denotes  the vectorization of matrix $B$ by stacking the columns (from left to right) of $B$  into a single column, and the asymptotic precision  matrix is
	\begin{equation}\label{eq:P(s)}
		P(s)=\Omega^{-1}\otimes (q^{-1} \Gamma(s)),
	\end{equation}
	where $\otimes$ denotes the Kronecker product and $\Omega$ is the covariance matrix  defined in (\ref{eq:e_t cov}).
\end{theorem}
An extension of Theorem \ref{Thm:asymp_normality} incorporating auxiliary estimates for the means and the sampling function is stated in Theorem \ref{Thm:asymp_aux} below.

It is now natural to consider the optimization of  the precision (information) matrix $P(s)$ under the constraint that the sampling rate $q=\E[s(\mbf{x}_t)]$ is fixed at a value in the interval $(0,1)$. In general,
one cannot expect to optimize $P(s)$ with respect to the Loewner order, namely, there is no    $P(s)$ which is optimal  in every   direction (e.g., \citet[Chapter 4]{pukelsheim:1993:optimal}). Instead, one may consider the optimization of a suitable scalar function of $P(s)$,  and most popularly, the determinant, which leads to the so-called D-optimality (e.g., \citet[Chapter 9]{pukelsheim:1993:optimal}).  By a property of Kronecker product,
$\det(P(s))$ is proportional to $\det(\Gamma(s))^{K}$, and hence we need to maximize $\det(\Gamma(s))$.

\begin{theorem}[D-optimality]
\label{Thm:D-opt}

	If the distribution of $\mbf{x}_t$ is elliptical as specified by \eqref{eq:x EC}, then  
	the following constrained optimization problem:
	\begin{equation}\label{eq:opt}
		\argmax_s \, \det( P(s) ) =\argmax_s \, \det( \Gamma(s) ) \quad \text{subject to}\quad \E[s(\mbf{x}_t)]=q\in (0,1],
	\end{equation}
	where the maximization is over all measurable $s:\bb{R}^p\rightarrow [0,1]$, has  solution
	\begin{equation}\label{eq:sample fun lev}
		s(\mbf{x})=1_{\{ \mbf{x}' \Sigma^{-1} \mbf{x} >r\}},
	\end{equation}
	where $r>0$ is chosen so that 
	\begin{equation}\label{eq:threshold quantile}
		pr( \mbf{x}'_t \Sigma^{-1} \mbf{x}_t >r)=q.
	\end{equation}
	This solution is almost everywhere unique with respect to the distribution of $\mbf{x}_t$.
\end{theorem}
Theorem \ref{Thm:D-opt} is a special case of Theorem \ref{Thm:D-opt relax} below, which is provided in Section \ref{Sec:relax}. 
We note that the optimality problem in Theorem \ref{Thm:D-opt} can be regarded as a special case of the general optimality problem formulated in \citet{pronzato:2006:sequential,pronzato:2020:sequential}.  Instead of obtaining an explicit solution as in \eqref{eq:sample fun lev}, the aforementioned studies focused on stochastic approximation algorithms and were in general constrained to the setup of i.i.d.\ trials. The explicit solution in Theorem \ref{Thm:D-opt}, although obtained under a more restrictive framework compared to \cite{pronzato:2006:sequential,pronzato:2020:sequential}, enables one to formulate explicit and efficient online sampling algorithms which can   be applied in the streaming time series context.

The tail quantile $r$   in \eqref{eq:threshold quantile} may not be unique if $pr( \mbf{x}'_t \Sigma^{-1} \mbf{x}_t >r)$ is not strictly decreasing with respect to $r$ (until reaching zero). One may eliminate this non-uniqueness by assuming that  the density  of $\nu$ in \eqref{eq:x EC} is positive.

Besides  D-optimality,  one may   consider some other  optimality criteria, e.g., A-optimality which minimizes $\tr (P(s)^{-1})$, E-optimality which maximizes the minimum eigenvalue of $P(s)$, T-optimality which maximizes $\tr(P(s))$, etc (e.g., \citet[Chapter 9]{pukelsheim:1993:optimal}).  Different criteria may lead to different optimal solutions of the sampling function $s$. We have chosen  D-optimality since the   solution has  intuitively appealing interpretation as well as computational advantage.

Theorem \ref{Thm:D-opt} leads to the following  simple deterministic sampling rule: $(\mbf{y}_t;\mbf{x}_t)$ is selected if   $\mbf{x}'_t \Sigma^{-1} \mbf{x}_t$ exceeds the threshold $r$. Such a sampling rule was proposed in   \cite{xie:2019:online} for the special case of Gaussian vector autoregression  model,  and was termed \emph{leverage score sampling (LSS)} since  
\[
	\ell_t=  \mbf{x}_t' \Sigma^{-1} \mbf{x}_t ,
\]
can be interpreted as a \emph{leverage score} of the  $\mbf{x}_t$ which marks the influence of this covariate point (cf.\ \citet[Section 10.2]{seber:2012:linear}).  In  \cite{xie:2019:online} the superiority of LSS   over a Bernoulli sampler was established under the same sampling rate $q$ in terms of the Loewner order.  Theorem \ref{Thm:D-opt} above  takes a further step to show that  under the D-optimality criterion, LSS is the best choice among the $\cl{S}(s)$ class in Definition \ref{Def:S} which includes the Bernoulli sampler.

To implement LSS in practice, one needs auxiliary  estimations of the means of $\mbf{y}_t$ and $\mbf{x}_t$ (recall that we have assumed centering \eqref{eq:centering}), the calculation of leverage score $\ell_t$ as well as the $q$-quantile of $\ell_t$. These issues will be discussed in a more general context  in Section \ref{Sec:aux} below.

\subsection{Relaxed Leverage Score Sampling (Relaxed-LSS)}\label{Sec:relax}

The hard thresholding found in \eqref{eq:sample fun lev}  employed in LSS selects exclusively   high-leverage  $\mbf{x}_t$ points.  There are   two potential risks of removing low-leverage design points in a linear regression: (a) if the linear regression relationship between the response and the covariates fails to hold in the low-leverage design region, then one would not be able to detect this departure;   (b) relying on high-leverage points  may make the regression particularly susceptible to the influence of outliers. Recall that the influence of an outlier is a combined effect of  the magnitude of   the regression residual  and the leverage of the design point, as measured by Cook's distance  \citep{cook:1977:detection} for instance. 

These potential problems suggest us to relax LSS by including a reasonable fraction of low-leverage points. This motivates the following  optimization modified from that in Theorem \ref{Thm:D-opt}:
\begin{equation}\label{eq:relax LSS opt}
	\argmax_s \, \det( \Gamma(s) ) \quad \text{subject to}\quad \E[s(\mbf{x}_t)]=q\in (0,1],\quad s(\mbf{x})\ge q_0, 
\end{equation}
where $s$ is the sampling function in Definition \ref{Def:S}, and $q_0\in [0,q]$. In other words, we include the additional constraint that there is  a  base sampling rate $q_0$ regardless of the design point $\mbf{x}_t$.  
\begin{theorem}\label{Thm:D-opt relax}
	Under the same conditions as Theorem \ref{Thm:D-opt},   the optimization  problem  \eqref{eq:relax LSS opt} has   solution 
	\begin{equation}\label{eq:s tilde}
		s(\mbf{x})=q_0+(1-q_0)1_{\{ \mbf{x}' \Sigma^{-1} \mbf{x} >r\}},
	\end{equation}
	where $r$ is chosen so that
	\begin{equation}\label{eq:threshold quantile relax}
		pr( \mbf{x}'_t \Sigma^{-1} \mbf{x}_t >r)=\frac{q-q_0}{1-q_0}.
	\end{equation}
	This  solution is almost everywhere unique with respect to the distribution of $\mbf{x}_t$.
\end{theorem}
Theorem \ref{Thm:D-opt relax} includes Theorem \ref{Thm:D-opt} as a special case  when $q_0=0$. 
This solution has the following interpretation:  at each time point $t$, with a small probability $q_0$, the sample unit $(\mbf{y}_t;\mbf{x}_t)$ will be selected regardless of the leverage score $\ell_t$ of $\mbf{x}_t$;  with a large probability $1-q_0$,  the hard thresholding of $\ell_t$ is applied to select sample. Such a strategy is also called rejection control in Monte Carlo computing \citep{liu2004monte}.
With a fractional of small-leverage points included, one may perform model diagnostics, e.g., using regression residuals,   to assess the goodness of fit \cite[Chapter 13]{lutkepohl:2005:new} over low-leverage design region.

\section{Auxiliary Estimates, Algorithm and Practical Implementation}\label{Sec:aux}

\subsection{Practical Implementation and Computational Complexity}

The Algorithm \ref{Alg:LSS} summarizes the relaxed-LSS-assisted online estimation procedure involving some of the  ingredients discussed below. 
\smallskip

\noindent\underline{\textbf{Incremental estimation of means}}

In practice, the means $\mbf{\mu}_Y=\E[\mbf{y}_t]$ and $\mbf{\mu}_X=\E[\mbf{x}_t]$ are often not zero and unknown.
Since the computational cost of updating  the means $\mbf{\mu}_X$ and $\mbf{\mu}_Y$ is minimal and having them estimated accurately is important for ensuring the consistency of $\wh{B}_{n,I}$,   we propose to update them  frequently, even at every step.
We suggest starting the LSS-assisted online estimation using initialized means based on a pilot stream sample, and then have them updated incrementally. 
\smallskip

\noindent\underline{\textbf{Real-time calculation of leverage score}}

An effective real-time calculation of leverage score requires less computational effort than updating the least square. In practice, the inverse covariance matrix $\Sigma^{-1}$ is unknown. 
To gain the computational advantage,   a real-time calculation of leverage score approach is used by incorporating a crude estimate of precision matrix.
A crude estimate of $\Sigma^{-1}$ will  affect only the efficiency of the leverage score sampling but not the consistency of $\wh{B}_{n,I}$.
We hence propose to use pilot estimation $\mathrm{P}_0:=\wh{\Sigma}_{n_0}^{-1}$ based on pilot data of size $n_0$ as a crude estimation of leverage score, i.e.
\begin{equation}
    \label{eq:ell_t}
    	\ell_t \approx  \mbf{x}_t' \mathrm{P}_0 \mbf{x}_t,
\end{equation}
so that we can update the leverage score with one vector-matrix multiplication, which has cost $O(p^2)$ per time point; or, alternatively, we update $\mathrm{P}_0$ sparsely in time, which has total cost $O(cp^2)$ with $c\ll n$, given $n$ as the total length of the observed stream size.
\smallskip

\noindent\underline{\textbf{Computational Complexity}}

The running time for the LSS-assisted online estimation depends on both the time to calculate the leverage scores and the time to update the model estimation using  sampled data. For an observed stream of size $n$, calculating leverage scores requires total $O(cp^2)$ time, and updating least squares estimation for sampled data requires  $O\left((qn)p^2\right)$ time with sampling rate $q\ll1$. So the total computational complexity of Algorithm \ref{Alg:LSS} is $O\left((qn+c)p^2\right)$, where $(qn+c)\ll n$.
The computational complexity of LSS or relaxed-LSS sampling methods hence is lower than the recursive least squares methods, where the latter needs to update the least squares estimates at every time point resulting in total $O(np^2)$ time. 
For the Bernoulli sampler (i.e., sampling function $s(\mbf{x})\equiv q$), the running time    is trivial. 
To gain a further computational advantage in practice, one may  use an efficient approximate computation of the leverage scores to perform relaxed-LSS algorithm (cf.\ \cite{ma:2015:statistical}).  The   approximation error analysis of the leverage scores can be found in \cite{drineas2012fast} and \cite{gittens2013revisiting}.
\smallskip

\noindent\underline{\textbf{Determination of the threshold} $r$}

Another important practical issue is the determination of the threshold $r$.   If  each $\mbf{x}_t$ is  Gaussian,  it can be shown that the sampling rate
\begin{equation}\label{eq:Q(m,r)}
	pr(\ell_t>r)=pr(\chi_p^2 >r),
\end{equation}
where $\chi_p^2$ denotes a Chi-square distribution with $p$ degrees of freedom. Hence for a predetermined sampling rate $q\in (0,1)$, one can choose $r$ based on \eqref{eq:Q(m,r)}. In reality, data often exhibit heavier-tailed fluctuation compared to Gaussian.  So one may start the sampling using a threshold $r$   determined by \eqref{eq:Q(m,r)}, and then replace it with the empirical quantile of $\ell_t$ observed thus far.
It is also possible to perform a pre-estimation of the tail quantile $r$ based on a pilot sample of $\mbf{x}_t' \Sigma^{-1} \mbf{x}_t$  (with $\Sigma^{-1}$ also pre-estimated). 

\subsection{Auxiliary Estimates}
Below we formulate a general asymptotic result extending Theorem \ref{Thm:asymp_normality} which incorporates   auxiliary estimates.   Suppose the assumptions in Section \ref{Sec:linear sys} hold, but now we do not assume \eqref{eq:centering}. 
 We shall suppose that for some $\alpha>2$,
\begin{align}\label{eq:alpha moment}
	\E\|\mbf{x}_t\|^\alpha<\infty, \quad t\in\bb{Z}_+, 
\end{align}
as well as
\begin{align}\label{eq:mu_X mean sq cons}
	\sup_{t } t^{1/2}  [\E \| \wh{\mbf{\mu}}_{t,X} -\mbf{\mu}_{X}\|^{\alpha}]^{1/\alpha} <\infty,~  \sup_{t } t^{1/2}  [\E \| \wh{\mbf{\mu}}_{t,Y} -\mbf{\mu}_{Y}\|^\alpha]^{1/\alpha} <\infty.
\end{align}       
The latter two relations are  moderate strengthening of the usual root-$n$ consistency.  If $\wh{\mbf{\mu}}_{t,X}$ and $\wh{\mbf{\mu}}_{t,Y}$ are sample means up to time $t$  and $\E \|\mbf{x}_t\|^{\alpha}<\infty$, $\E \|\mbf{y}_t\|^{\alpha}<\infty$, then \eqref{eq:mu_X mean sq cons} holds if a strong mixing condition  holds for $(\mbf{x}_t)$ and $(\mbf{y}_t)$ \citep{yokoyama:1980:moment}, or if they are short-range dependent linear processes \cite[Proposition 4.4.3]{surgailis:2012:large}.

Next, we assume that there is a 
family of sampling functions $s=s_q$ indexed by the sampling rate $q$, which satisfies
\begin{align}\label{eq:sampling fun cond aux}
	\lim_{q\rightarrow 0}\frac{q^{1-2/\alpha}\lambda_{\max}(\Gamma(s))^{1/2}}{\lambda_{\min}(\Gamma(s))}=0,
\end{align}
where $q=\E[s(\mbf{x}_t)]$ is the sampling rate and $\lambda_{\max}(\Gamma(s))$ and $\lambda_{\min}(\Gamma(s))$ stand for the largest and the smallest eigenvalue of $\Gamma(s)$ respectively. Condition \eqref{eq:sampling fun cond aux}, roughly speaking,   ensures that the magnitude of $\Gamma(s)$ decays slower than a power of $q$ as $q\rightarrow 0$. This will be verified for the sampling function corresponding to the relaxed-LSS method in Lemma \ref{Lem:aux} below.

We   suppose that
$\wh{s}_t\in [0,1]$ is an estimate of the  plugged-in sampling function   $s(\mbf{x}_t-\mbf{\mu}_X)$ based on the data stream observed up to time $t$ such that as $t\rightarrow\infty$ ($q$ fixed),
\begin{align}\label{eq:s consistent}
	\wh{s}_t- s(\mbf{x}_t-\mbf{\mu}_X) \ConvP 0.
\end{align}
This consistency condition will also be verified in Lemma \ref{Lem:aux} below for the auxiliary estimates involved in the relaxed-LSS method. 
\begin{theorem}\label{Thm:asymp_aux} 
	Suppose that the conditions  \eqref{eq:alpha moment}, \eqref{eq:mu_X mean sq cons}, \eqref{eq:sampling fun cond aux} and \eqref{eq:s consistent} hold. Let the   estimator of $B$ based on stream up to time $n$ be as 
	\[
	\wh{B}_{n,s}=\left( \sum_{t=1}^n  \wt{\mbf{x}}_t   \wt{\mbf{x}}_t'   1_{\{U_t\le   \wh{s}_t \}} \right)^{-1} \left( \sum_{t=1}^n \wt{\mbf{x}}_t \wt{\mbf{y}}_t'    1_{\{U_t\le   \wh{s}_t \}} \right),
	\]
	where $U_t$'s are as in Definition \ref{Def:S},  and $\wt{\mbf{x}}_t=\mbf{x}_t-\wh{\mbf{\mu}}_{t,X},\quad\wt{\mbf{y}}_t=\mbf{y}_t-\wh{\mbf{\mu}}_{t,Y}.$  Then  we have the decomposition
	\begin{equation}\label{eq:asymp norm lev aux} 
		\sqrt{N}( \wh{B}_{n,s}   -  B)  =  M_n+R_n,
	\end{equation}
	where as  the total stream size $n\rightarrow\infty$,
	\begin{equation}\label{eq:asymp norm M_n}
		\vect(M_n) \ConvD \cl{N}(\mbf{0},  P(s)^{-1} )
	\end{equation}
	as $n\rightarrow\infty$ with    $P(s)$   as in \eqref{eq:P(s)} but with $\Gamma(s)$ in \eqref{eq:Gamma(s)} redefined as
	\begin{equation}\label{eq:Gamma(s) center}
		\Gamma(s)=\E\left[ s(\mbf{x}_t-\mbf{\mu}_X) (\mbf{x}_t-\mbf{\mu}_X) (\mbf{x}_t-\mbf{\mu}_X)' \right],
	\end{equation} 
	which we assume to be non-singular.
	The term $R_n$ satisfies for any $\delta>0$,
	\begin{equation}\label{eq:R_n neg}
		\lim_{q\rightarrow 0} \limsup_n pr(\|P(s)^{1/2}\vect(R_n)\|>\delta)= 0.
	\end{equation}
\end{theorem}

The double limit in \eqref{eq:R_n neg} says that when  the sampling rate is small, as typically desired in practice, the term $R_n$ is  negligible compared to $M_n$.  Note that the same double limit with $R_n$ replaced by $M_n$ will not be zero due to \eqref{eq:asymp norm M_n}. 


\begin{algorithm} 
\caption{Relaxed-LSS-Assisted Online Estimation of Stationary Linear System}\label{Alg:LSS}
	\textbf{Initialization}:\\
	Choose a sampling rate $q\in (0,1)$\;
	Choose a base sampling rate $q_0<q$\;
	Initialize the estimates of $\mbf{\mu}_X$, $\mbf{\mu}_Y$, $\Sigma^{-1}$ and $B$ based on a   pilot sample\;	
	\smallskip	
	\textbf{Online Estimation}:\\ 
    \While{New sample $(\mbf{y}_t;\mbf{x}_t)$ at time $t$ arrives}{		
		Update     $\mbf{\mu}_X$ and $\mbf{\mu}_Y$\;		
		Calculate $\ell_t$ based on $\mathrm{P}_0$ in \eqref{eq:ell_t}\;
  \eIf{Bernoulli($q_0$) random number = 1}{
    Update the  estimates  of $B$ and $\Omega$  with new centered sample  $(\mbf{y}_t-\mbf{\mu}_Y;\mbf{x}_t-\mbf{\mu}_X)$\;
  }{
      \If{$\ell_t > r$  }{
      Update the  estimates  of $B$ and $\Omega$  with new centered sample  $(\mbf{y}_t-\mbf{\mu}_Y;\mbf{x}_t-\mbf{\mu}_X)$\;
      }
  }
  Update $r$ based on \eqref{r_estimate} (or use \eqref{eq:Q(m,r)} when the sample size is small). 
}	
\end{algorithm}

The following lemma shows that the conditions \eqref{eq:sampling fun cond aux} and \eqref{eq:s consistent} are satisfied in the context of relaxed-LSS, and hence justify   the procedure  in Algorithm~\ref{Alg:LSS}. 
\begin{lemma}\label{Lem:aux}
	~
	\begin{enumerate}[(a)]
		\item Suppose $s(\mbf{x})=s_q(\mbf{x})=q_0+(1-q_0) 1_{\{\mbf{x}'\Sigma^{-1}\mbf{x}>r\}}$ ($q_0$ and $r$ depend on $q$) is the sampling function of the relaxed-LSS  as in Theorem \ref{Thm:D-opt relax}, where $q_0=q_0(q)$ satisfies that  for some constant  $c\in (0,1)$  
		\begin{equation}\label{eq:q_0 restr}
			q_0\le c q.
		\end{equation}
		When $\alpha\in (2,4]$,  assume in addition  that for some constant $c>0$ and $\beta\in (\alpha, 4\alpha/(4-\alpha) )$ (right boundary understood as $+\infty$ when $\alpha=4$), we have
		\begin{align}\label{eq:tail lower bound}
			\nu(x,\infty)> c x^{-\beta}
		\end{align}
		for all sufficient large $x$, where $\nu$ is as in \eqref{eq:x EC}.    Then the condition \eqref{eq:sampling fun cond aux} holds.
		\item 
		Suppose that $\wh{\Sigma}_n^{-1}$ is a consistent estimate of $\Sigma^{-1}$ based on the stream observed up to time $n$, which can be realized by
		\begin{equation}\label{sigma inv est}
\wh{\Sigma}_n^{-1}=\left( \frac{1}{M_n}\sum_{t=1}^n J_t (\mbf{x}_t-\wh{\mbf{\mu}}_{t,X}) (\mbf{x}_t-\wh{\mbf{\mu}}_{t,X})'\right)^{-1},
\end{equation}
with a desirable small update rate $u\in (0,1)$, which corresponds to i.i.d.\ Bernoulli($u$) random variables $(J_t)$ independent of everything else, where  $\wh{\mbf{\mu}}_{t,X}$ and $\wh{\mbf{\mu}}_{t,Y}$   be  estimators of $\mbf{\mu}_X$,  $\mbf{\mu}_Y$  based on the stream observed up to time $t$, respectively.

		Define the leverage score incorporating the auxiliary estimates as
		\begin{align}\label{eq:tilde ell}
			\wt{\ell}_t=\wt{\mbf{x}}_t'  \wh{\Sigma}_t^{-1}  \wt{\mbf{x}}_t.
		\end{align}
			Let $q_0$ be the base sampling rate for relaxed-LSS in Section \ref{Sec:relax}.
		Suppose that
		$\wh{r}_n$ is a consistent estimate of 
		\begin{equation}
		r(q,q_0):= \inf\left\{r\ge 0:\  pr((\mbf{x}_t-\mbf{\mu}_{X})'\Sigma^{-1}(\mbf{x}_t-\mbf{\mu}_{X})\le r) \ge  \frac{q-q_0}{1-q_0}\right\}
		\end{equation}  based on the stream observed up to time $n$, which can be realized by  
\begin{equation}\label{r_estimate}
\wh{r}_n=\inf \left\{r\ge 0: \frac{1}{n}\sum_{t=1}^n 1_{\{\wt{\ell}_t\le r\}}\ge \frac{q-q_0}{1-q_0}\right\}.
\end{equation}  Let the estimated plugged-in sample function be
		\begin{align}\label{eq:s hat t}
			\wh{s}_t = q_0+(1-q_0)1_{\{ \wt{\ell}_t> \wh{r}_t\}}.
		\end{align}
	 Then the condition \eqref{eq:s consistent} holds. 
	\end{enumerate}
\end{lemma}
Examining the proof reveals that   the condition \eqref{eq:q_0 restr} can be   relaxed by allowing a power of $q$. We omit such a generalization  here for simplicity. We also note that the assumption \eqref{eq:tail lower bound} is not stringent. In particular, suppose the tail probability $\nu(x,\infty)$ is regularly varying  with index $-\gamma$ (see \cite{bingham:1989:regular}, which roughly speaking,  says $\nu(x,\infty)$ behaves like $x^{-\gamma}$)  as     $x\rightarrow \infty$, $\gamma>2$. This regular variation assumption includes   common heavy-tailed distributions    such as Pareto distributions and $t$-distributions. Then   in view of Potter's bound \cite[Theorem 1.5.6]{bingham:1989:regular}, one can find $\alpha$ and $\beta$ satisfying $2<\alpha<\gamma<\beta<4\alpha/(4-\alpha)$  so that  the conditions  \eqref{eq:alpha moment} and \eqref{eq:tail lower bound} both hold.

\section{The Open Power System Data Application}\label{Sec:data}


In this section, we apply our LSS and relaxed-LSS methods and benchmark Bernoulli method to the~\cite{realdata} for real-time inference on the dynamics of power grid load profiles and one-step ahead  prediction. 
We compare the proposed LSS-based methods to the benchmark sampling method and ``full sample'' estimation and demonstrate the strengths of real-time estimation and prediction of the proposed methods.

\subsection{Open Power System Data}
Geographically, the Open Power System Data consist $37$ European countries that cover European Union and neighboring countries. The data measures the total load (in 
Terawatt hour, TWh) for a country, control area or bidding zone. The total load is a power statistic, which is defined as, roughly speaking, the total power generated or imported minus the power being consumed at power plant, stored or exported. More specifically, the data reported by ENTSO-E Transparency Platform are used due to its high efficiency in data reporting, which results in a subset of 19 countries. 
 The selected multivariate time series streams are recorded from 2006-01-01, 00:00:00 Coordinated Universal Time (UTC) to 2018-12-14, 23:00:00 UTC.
There are total $113,544$ time points been observed in the $19$-dimensional electricity load stream, which are complete without missing values.

\begin{figure}[h]
	\begin{center}
		\includegraphics[width=4.5in]{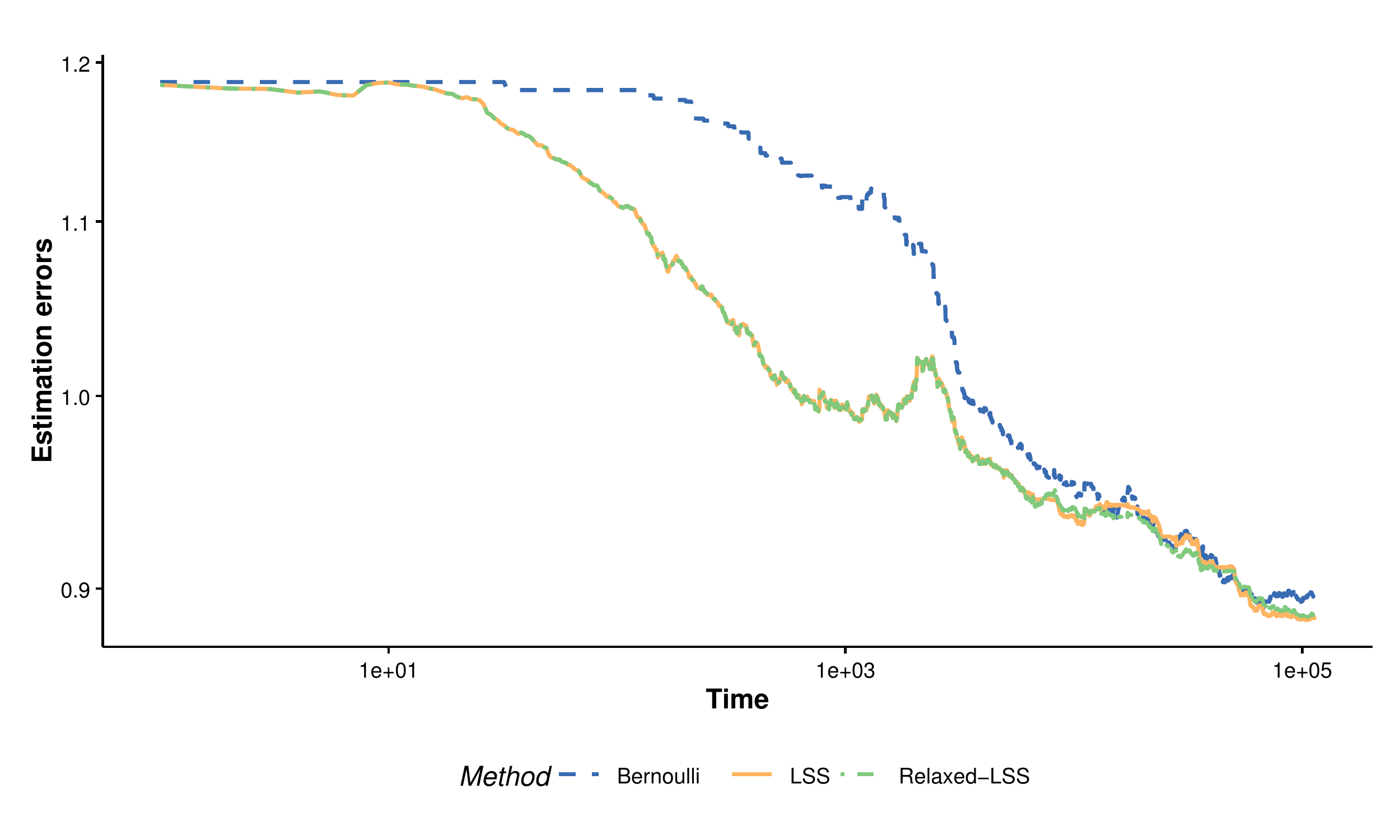}
		\vspace{-0.5cm}
		\caption{Comparison of estimation errors for each of the time points of power load data.}\label{fig:realdata_estimation}
	\end{center}
\end{figure}

\subsection{Seasonal VARX modeling and accuracy measurements}
Electricity loads exhibit strong seasonality compared to loads one day (24 hours) earlier since it was aggregated to hourly temporal resolution.  
We consider the seasonal vector autoregression model with centering:
\begin{equation}\label{eq:sVARX}
	\mbf{y}_t  = \sum_{i=1}^{p_1}\Phi_i \mbf{y}_{t-i}  +\sum_{i=1}^{p_2}\Theta_i \mbf{y}_{t-24i} + \mbf{e}_t, 
\end{equation} 
where we choose the model order as $p_1=2$, $p_2=1$ incorporating the daily seasonality. The weather, electricity prices, and clean energy generation and capacities can be added in to this seasonal VARX model as exogenous variables. However, due to the large portion and complicated missing pattern in the database, we did not include them in the real-time analysis. Our goals are to estimate the model parameters $\Phi's$ and $\Theta's$ in real time, which represent the dynamic dependence of the electricity loads, and the one-hour ahead forecasting (prediction) of electricity loads, which is crucial for effective scheduling and energy management in power grids. 

We denote the $\hat{B}_{t}$ as the real time estimation of model parameter matrix $B$ at time point $t$, and $\hat{B}_{\text{Full}}$ is the offline model parameter estimation based on the entire dataset as a substitute of the unknown population $B$. We use the relative errors to measure the parameter matrix estimation accuracy $	|| \hat{B}_{t} -\hat{B}_{\text{Full}}||_{F}/|| \hat{B}_{\text{Full}}||_{F}$.

At time point $t$, if the sample unit $(\mbf{y}_t;\mbf{x}_t)$ is sampled, we will update the estimation $\hat{B}_{t}$ using least squares; otherwise the estimation from previous time point will be retained as the current estimation. The results are demonstrated in Figure~\ref{fig:realdata_estimation}, where the x-axis represents the time of updates.

\begin{figure}[ht!]
	\begin{center}
		\includegraphics[width=5.0in]{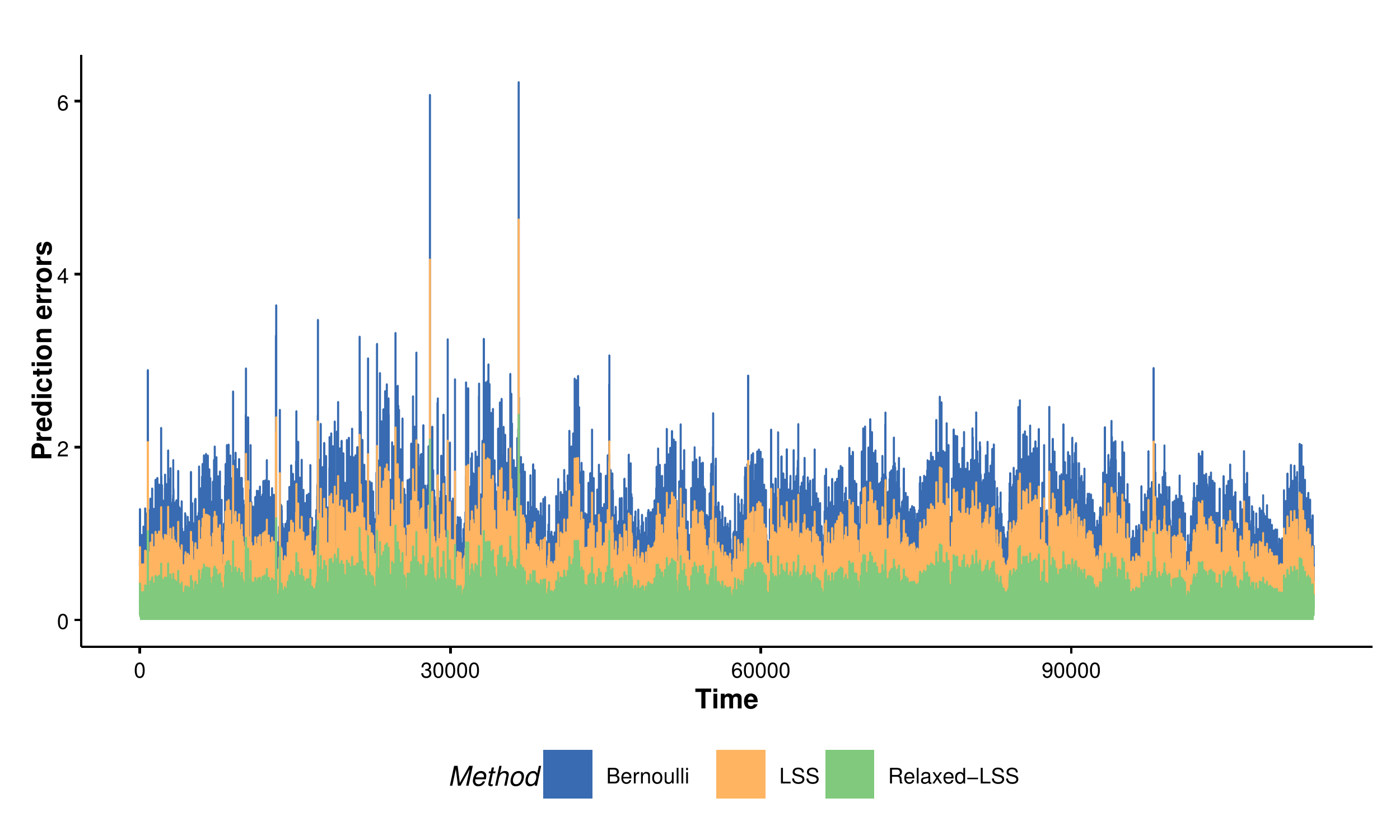}
			\vspace{-0.5cm}
		\caption{Comparison of one-step ahead prediction errors for each of the time point of power load data.}\label{fig:realdata_prediction}
	\end{center}
		\vspace{-0.5cm}
\end{figure}

For the pointwise prediction accuracy, we compute the one-step (one-hour) ahead relative prediction errors for each of the time points as:
$||\hat{\mbf y}_{t+1} - \mbf{y}_{t+1} ||/||\mbf{y}_{t+1} ||$.
We visualized the prediction errors in Figure~\ref{fig:realdata_prediction}, where x-axis represents hourly time.

\begin{figure}
     \centering
     \begin{subfigure}[b]{0.32\textwidth}
         \centering
         \includegraphics[width=\textwidth]{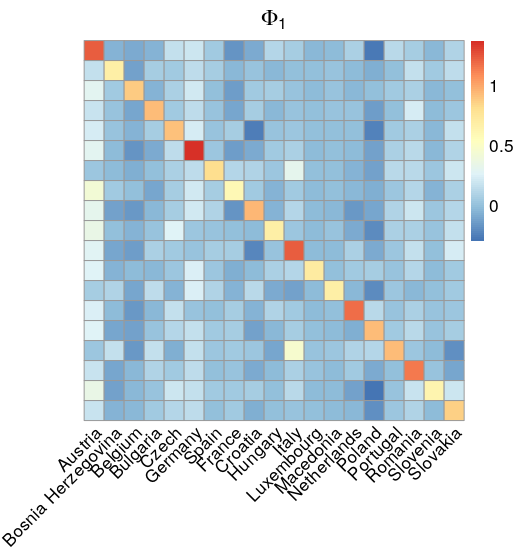}
     \end{subfigure}
     \hfill
     \begin{subfigure}[b]{0.32\textwidth}
         \centering
         \includegraphics[width=\textwidth]{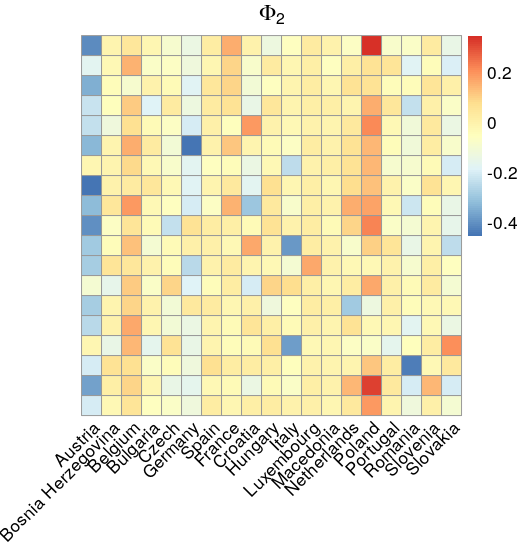}
     \end{subfigure}
     \hfill
     \begin{subfigure}[b]{0.32\textwidth}
         \centering
         \includegraphics[width=\textwidth]{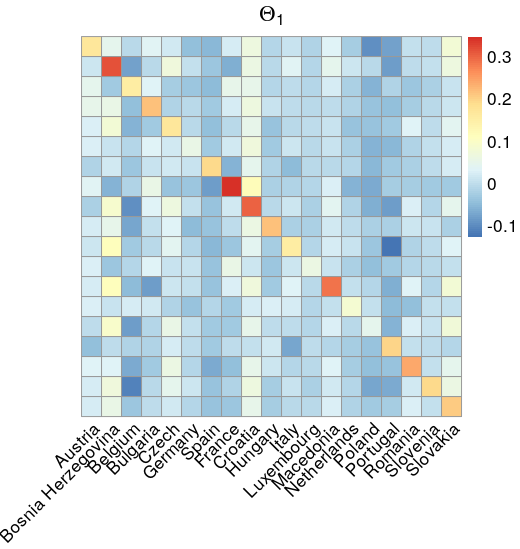}
     \end{subfigure}
        \caption{Visualization of the estimated coefficient matrices $\Phi_1$, $\Phi_2$ and $\Theta_1$ at time $t=113,424$ using relaxed-LSS method.}
        \label{fig:three graphs}
\end{figure}

\subsection{Results}
We compare our LSS and relaxed-LSS methods to the benchmark Bernoulli sampling method for parameter matrix estimation accuracy and prediction accuracy. 
For all three methods, we use the same pilot data of size $500$ to estimate the model order and initial values for model parameter $B$ and precision matrix $\mathrm{P}_0$.
The sampling rate is $q =0.05$ for all three methods, and the base sampling rate for relaxed-LSS is $q_0 =0.025$. The update rate for the inverse covariance matrix is $0.025$. We handle the data in a streaming fashion and take data samples and estimate the model in real time. 


As an illustration, Figure~\ref{fig:three graphs} depicts the estimated coefficient matrices $\Phi_1$, $\Phi_2$ and $\Theta_1$ at time $t = 113,424$, which is the last update, using relaxed-LSS method. 
The cross-correlations at lag 1 (coefficient matrix $\Phi_1$) between Italy and Portugal, Austria and France, as well as Italy and Spain are higher than others, which reflect their positive interdependence in electricity consumption.  
For the cross-correlations at lag 2 (coefficient matrix $\Phi_2$), Austria and several countries are negatively correlated, and  Poland and several countries are positively correlated with magnitudes  higher than others. We also notice that Germany has higher autocorrelation at lag 1 (positively)   and lag 2 (negatively). France, Bosnia Herzegovina, Croatia and Macedonia have higher daily-seasonal  correlation than other countries. It is also worth noting that those auto and cross correlations are dynamically evolving so that the Figure~\ref{fig:three graphs} is just a snapshot of the relationships at the time point. The computational cost of online estimation of those model coefficient matrices  could be higher if other variables of the power grid system are included in the model, such as electricity prices, wind and solar power generation and capacities. Those additional variables are available at different temporal resolutions or time frames on Open Power System Data platform. The need for real-time inference under complicated models and high computing costs reinforce the necessity of data reduction methods in analyzing IoT sensor streams.

In terms of estimation accuracy, LSS and relaxed-LSS methods behave similarly and outperform the Bernoulli sampling most of the time during updates. Overall, estimation errors for all three methods are decreasing along the update time, which suggests that the online estimates converge to the ``full sample'' estimate. However, in practice, one cannot afford to wait for the ``full sample'' due to the need for real-time monitoring in the power grid. Thus, given limited access to the data under computational constraint,  the faster it converges to the ``full sample'' estimate the better.  
Especially, given the same initial values, LSS and relaxed-LSS methods achieve better estimation accuracy at the early stage of updates than those of the Bernoulli method.   Eventually, LSS, relaxed-LSS and even Bernoulli sampling estimates are close to the ``full sample'' estimate since a large enough amount of samples are used for updating in all three sampling methods.

It is worth mentioning that there are a few sudden increases of estimation errors shortly after time at $1000$. We believe they are caused by the abnormal points observed in dimension 12 Luxembourg on 2011-03-27 at 22:00:00 UTC and in dimension 13 Macedonia on 2010-03-28 at 01:00:00 UTC. Those abnormal points were high leverage score points and thus were sampled by the LSS and relaxed-RSS methods. The corresponding model estimates deviated from the ``full sample'' estimates, which lead to sudden increases in estimation errors but were soon corrected by new data points. This phenomenon reflects the advantage of LSS-based sampling methods in capturing the influential or abnormal data points during stream monitoring. It is an important feature that LSS can be applied in the online monitoring of the dynamic dependence of the power grid system for security or online decision-making purposes. However, the phenomenon also reflects the limitation of the leverage based sampling method in lacking of robustness. The online estimation based on leverage score sampling is sensitive to the changes in underlying data generating processes or outliers. 

\begin{figure}[ht!]
	\begin{center}
		\includegraphics[width=5.0in]{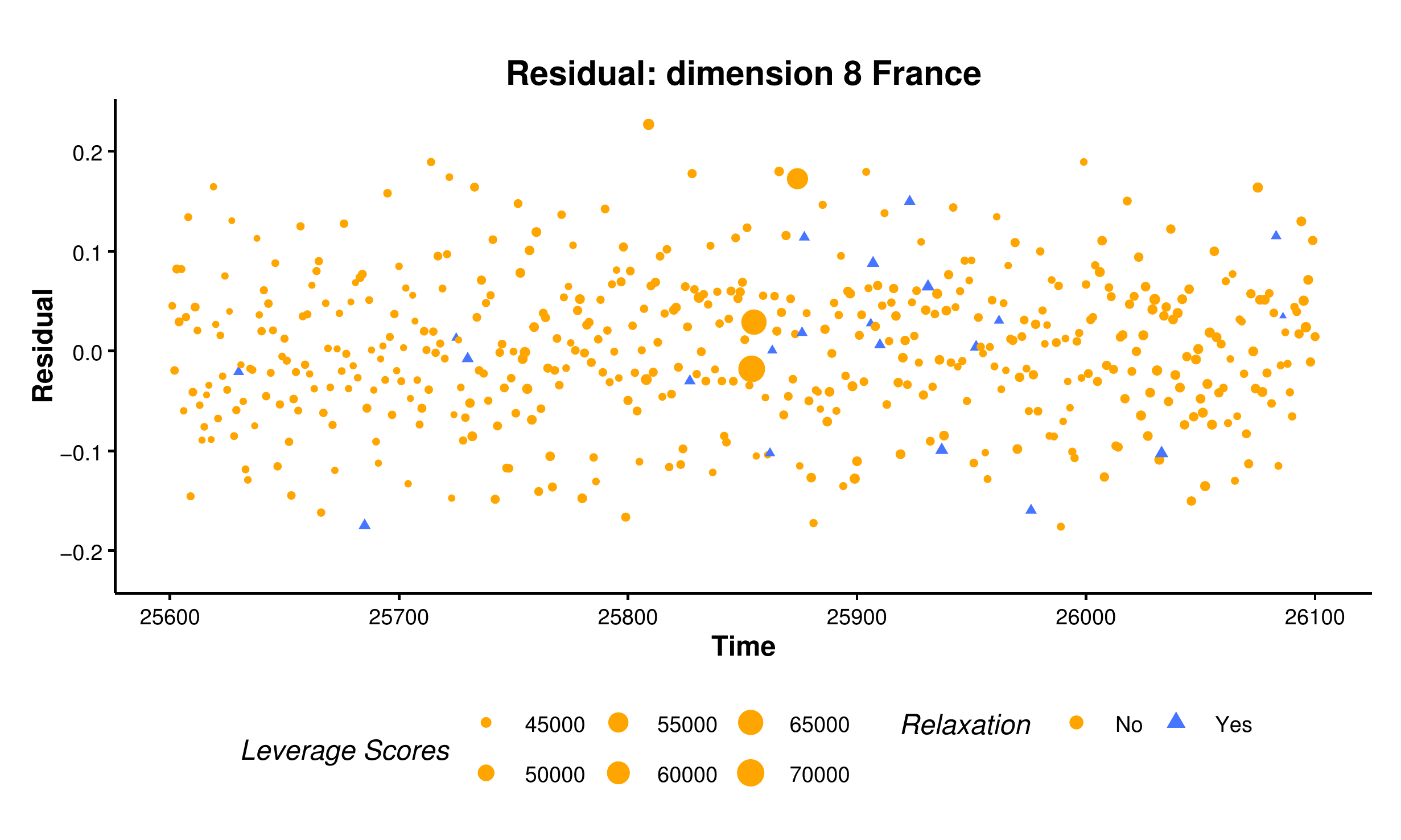}
			\vspace{-0.5cm}
		\caption{Residual plot  from regression fits for France, which is calculated based on the final estimate of relaxed-LSS along all historical data. A filled circle  denotes a point selected by the leverage score thresholding rule with the size of the filled circle indicating the leverage score, and a triangle point indicates a   point selected by the Bernoulli sampling rule.}\label{fig:realdata_diagnosis}
	\end{center}
	\vspace{-0.5cm}
\end{figure}

In addition to theoretically verified properties of LSS-based methods in parameter estimation, we also consider the prediction as a practical performance measurement. The accurate prediction of the electricity loads is crucial for power grid system in resource allocation and demand management. Prediction is also an important task for general IoT sensor data stream analysis due to similar reasons.
For prediction accuracy, the relaxed-LSS method consistently achieves the smallest prediction errors among the three compared methods; while Bernoulli method always delivers the largest prediction errors. The superiority of relaxed-LSS over LSS in prediction may be due to its inclusion of low-leverage covariate points. We also note the two peaks in prediction errors before and after time 30000, which are also caused by the abnormal points in Macedonia and Luxembourg.

Lastly, to examine the goodness of model fitting with relaxed-LSS sampling, we display the regression residuals from a segment of a selected sample of size $500$ from France in Figure~\ref{fig:realdata_diagnosis}, which are calculated based on the final estimate of relaxed-LSS on all historical data. In Figure~\ref{fig:realdata_diagnosis}, a filled circle denotes a point selected by the leverage score thresholding rule with the size of the filled circle indicating the leverage score, and a triangle indicates the low leverage score point. The residual plot shows that the linear relation proposed in the relaxed-LSS sampler holds equally well for both the points selected by a high leverage score and by {Bernoulli} sampling. 

\section{Simulation Studies for general applicability}\label{Sec:simulate} 
Although the proposed relaxed-LSS demonstrates better performance in inference and prediction on Open Power System Data, we also conduct simulation studies to 
evaluate the effectiveness of LSS and its relaxed version in general settings. We compare the LSS, relaxed-LSS and the Bernoulli sampling method, which corresponds to $s(\mbf{x})\equiv q$ in Definition \ref{Def:S}. We generate the multi-dimensional time series which follows the VARX model defined in~\eqref{eq:VARX} with $K=10$, $p_1=1$, and $p_2=1$. We consider the multivariate {\qcr Gaussian} and the multivariate {\qcr StudentT} distributions for noise process $({\mbf e}_t)$ and extraneous process $({\mbf v}_t)$.

\begin{figure}[ht!]
	\begin{center}
		\includegraphics[width=4.5in]{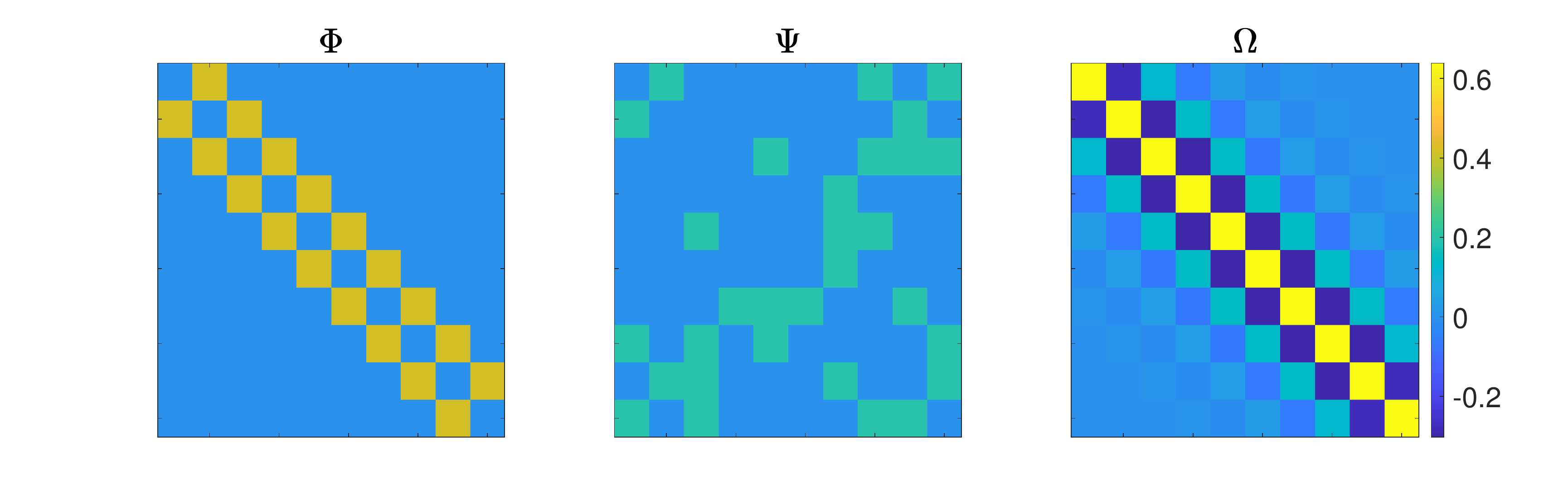}
		\caption{Visualization of the coefficient matrices $\Phi$ and $\Psi$, and covariance matrix for the error process $\Omega$.}\label{Fig:AR_matrix}
	\end{center}
\end{figure}

\subsection{Multivariate Gaussian Distribution Case}\label{sim:Gaussian}

\begin{figure}[ht!]
	\begin{center}
		\includegraphics[width=4.5in]{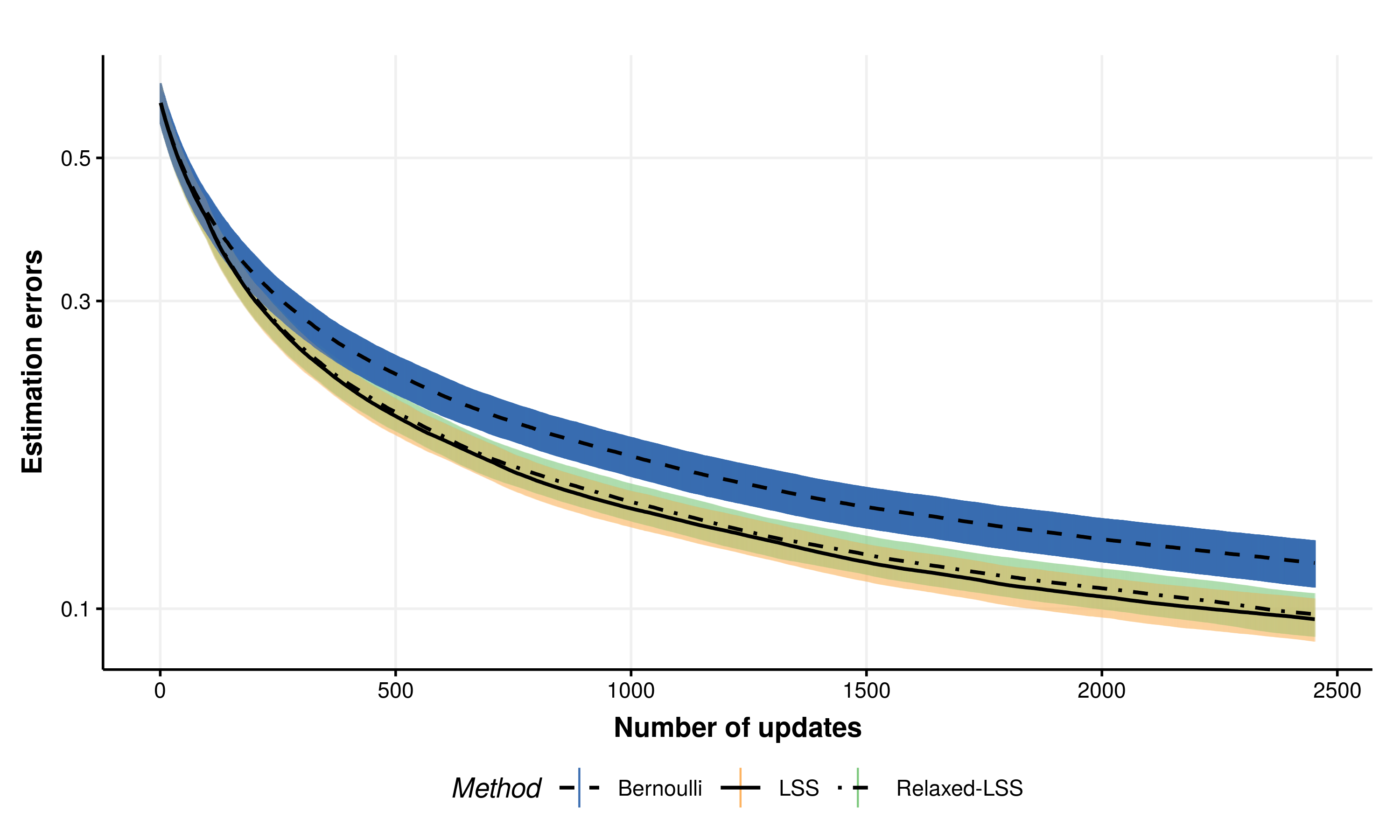}
		\caption{Comparison of estimation errors for simulated $10$ dimensional time series under {\qcr Gaussian} setting. Visualized for the first $2,500$ updates with underlying time series of length $25,000$. The average and one standard deviation error bar of the estimation errors are plotted over $1,000$ independent replicates.}\label{Fig:Gaussian}
	\end{center}
\end{figure}

We first consider the {\qcr Gaussian} case where the noise process $({\mbf e}_t)$ follows the i.i.d. multivariate Gaussian distribution $N({\mbf 0},\Omega)$ and the extraneous process $({\mbf v}_t)$ follows i.i.d. multivariate Gaussian distribution $N({\mbf 0},I_{Kp_2})$, where $I_{Kp_2} \in \bb R^{Kp_2}$ is the identity matrix.  We follow~\cite{qiu:2015:robust} to generate the coefficient matrices $\Phi$ and $\Psi$, and the covariance matrix for the error process $\Omega$, see Figure~\ref{Fig:AR_matrix} for visualization.

We generate the $10$-dimensional time series of length $n = 25,000$.
The sampling rate is $q =0.1$ and base sampling rate for relaxed-LSS is $q_0 =0.05$. The update rate for the inverse covariance matrix is $ 0.1$. A pilot sample of size $100$ is used to calculate initial estimations. We define the estimation error as $||\hat{B}_{\tau} -B||_{F}/||B||_{F}$,
where $\hat{B}_{\tau}$ is the $\tau$th update of the estimation of model parameter matrix $B$ (here we use $\tau$ to denote the number of updates to distinguish from the notation of time point $t$) and $||\cdot||_F$ denotes the Frobenius norm.


Figure~\ref{Fig:Gaussian} displays the average and one standard deviation error bar of the estimation errors at each update for each of the compared methods over $1,000$ independent replicates. 
Focusing on the first $2,500$ updates for each of the compared methods, we observe that LSS method achieve smallest estimation errors compared to relaxed-LSS and Bernoulli methods as predicted by our optimality theory. We note that the estimation errors for both LSS and relaxed-LSS are significantly smaller than those of Bernoulli method.



\subsection{Multivariate Student T Distribution Case}\label{sim:StudentT}

\begin{figure}[ht!]
	\begin{center}
		\includegraphics[width=4.5in]{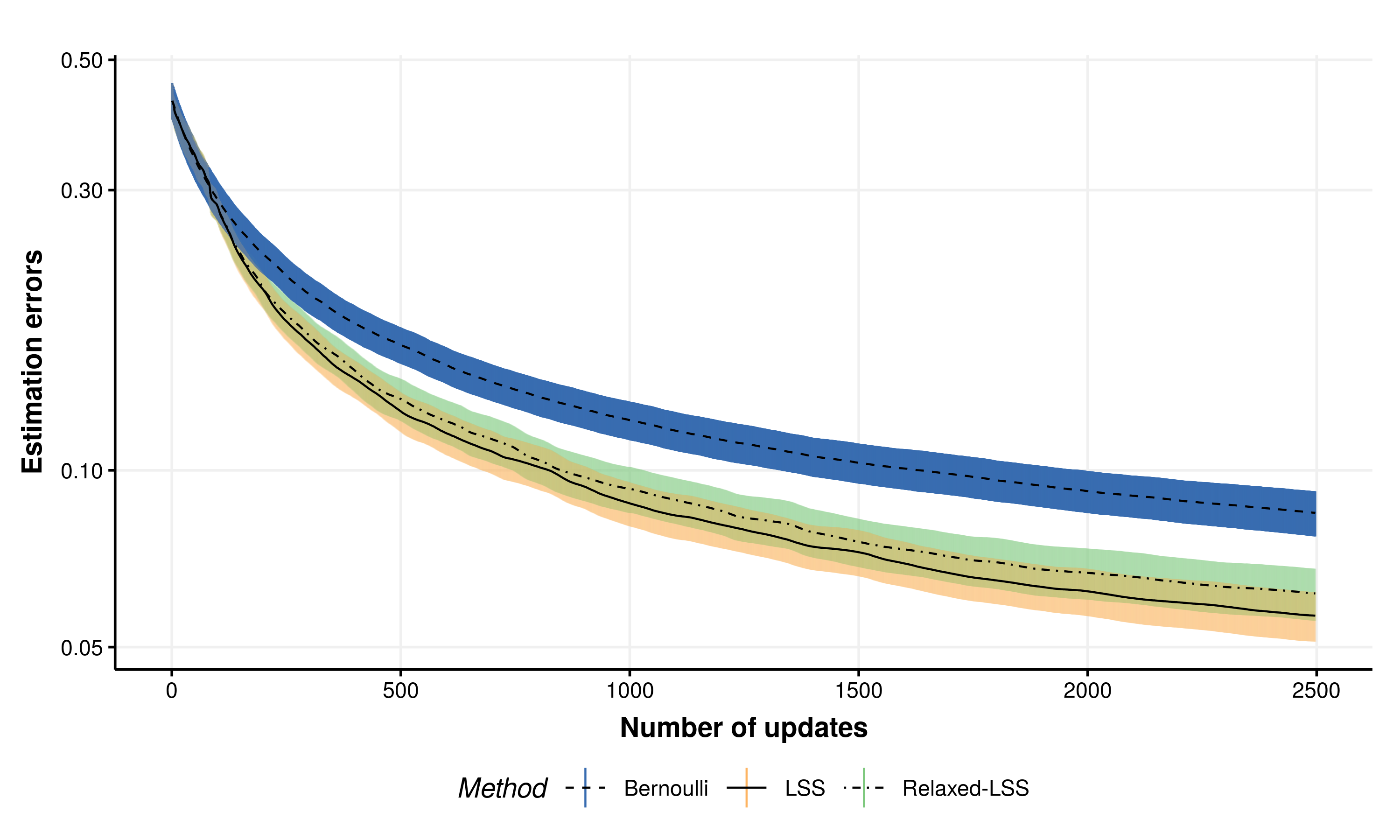}
		\caption{Comparison of estimation errors for simulated $10$ dimensional time series under {\qcr StudentT} setting. Visualized for the first $2,500$ updates with underlying time series of length $25,000$. The average and one standard deviation error bar of the estimation errors are plotted over $1,000$ independent replicates.}\label{Fig:StudentT}
	\end{center}
\end{figure}

We next consider the {\qcr StudentT} case.
The noise process $({\mbf e}_t)$ follows the i.i.d. multivariate student $t$ distribution with location $\bf 0$, degree of freedom $3$ and scale matrix $\Omega$, and the extraneous process $({\mbf v}_t)$ follows i.i.d. multivariate student $t$ distribution with location $\bf 0$, degree of freedom $3$ and scale matrix $I_{Kp_2}$.

We generate the time series following~\cite{remillard2012copula,qiu:2015:robust}.
Particularly, we first simulate an initial observation $\mbf{y}_1$, extraneous variables $\mbf{v}_1,\ldots,\mbf{v}_{t-1}$, and innovations $\mbf{e}_2,\ldots,\mbf{e}_t$. 
After generating  $(\mbf{y}_1',\mbf{e}_2',\ldots,\mbf{e}_t')'$ and $(\mbf{v}_1',\ldots,\mbf{v}_{t-1}')'$, one can form $(\mbf{y}_1',\mbf{y}_2',\ldots,\mbf{y}_t')'$ recursively following the iterative algorithm in \cite{remillard2012copula}. The model coefficient matrices $\Phi$ and $\Psi$, and the covariance matrix $\Omega$ are the same as the {\qcr Gaussian} case. 
We generate the multi-dimensional time series of length $t = 25,000$ and set the sampling rate $q =0.1$, the base sampling rate $q_0 =0.05$ and the update rate for the inverse covariance matrix is $ 0.1$. The pilot sample size is $100$.

Figure~\ref{Fig:StudentT} presents the average and one standard deviation error bar of estimation errors against number of estimate updates, which are based on $1,000$ independent replicates.
Similar to the {\qcr Gaussian} case, the LSS method achieves the best performance as predicted by our optimality theory; the LSS and relaxed-LSS methods are comparable;   both methods are significantly better than the Bernoulli method in terms of estimation errors, where the wider difference compared to the Gaussian case is due to the fact that the $t$-distribution has a heavier tail and hence generates a larger leverage effect.

\section{Conclusion}\label{con}
We introduced a class of online sample selection (sampling) methods for large scale streaming time series with application in the online analysis of electricity power grid data. We provide a solution to online statistical inference of high speed multidimensional time series streams under computational constraint.  The proposed methods were motivated by optimal designs in design of experiments and were applied to the high temporal resolution data streams in power grid system as an example of IoT sensor network data stream analysis. 
The proposed methods were based on a relaxed version of leverage score sampling and achieved an optimality criterion. Therefore, the proposed methods enjoyed the optimality in online sampling theoretically and improved the computational efficiency of the online analysis.
The elliptical distributed synthetic data and electricity consumption real data analysis demonstrated the effectiveness of the proposed sampling methods.  
Our proposed relaxed-LSS method provides online analysis of electricity loads through data selection without loss of identification of electricity consumption patterns and flexibility.
Our work was based on the stationary linear multivariate time series models for streaming data modeling, which serves as a foundation for tackling the more involved non-stationary case. We shall leave the study of sampling of non-stationary data to future work. 


\begin{acks}[Acknowledgments]
We thank the Editor, Associate Editor, and two anonymous reviewers for many valuable comments and suggestions.
\end{acks}

\begin{funding}

This work is partially supported by NIH awards R01MD018025, R03AG069799,  NSF awards DMS-1903226, DMS-1925066 and DMS-2124493.

\end{funding}




\bibliographystyle{chicago}
\bibliography{Bib}

\end{document}